\documentclass[10pt, conference, compsocconf]{IEEEtran}
\ifCLASSINFOpdf
\else
\fi
\usepackage{graphicx}
\usepackage{amsfonts}
\usepackage{amsmath}
\usepackage{multirow}
\usepackage{caption}
\usepackage{float} 
\usepackage{subfigure}

\begin{document}
%
\title{Pedestrian Spatio-Temporal Information Fusion For Video Anomaly Detection}


\author{\IEEEauthorblockN{1st Chao Hu\textsuperscript{*}}
\IEEEauthorblockA{AI Lab\\
Unicom (Shanghai) Industry Internet Co., Ltd.\\
Shanghai, China\\
huchao.000@gmail.com}
\and
\IEEEauthorblockN{2nd Liqiang Zhu}
\IEEEauthorblockA{AI Lab\\
Unicom (Shanghai) Industry Internet Co., Ltd.\\
Shanghai, China\\
zhulq0@gmail.com}
}

\maketitle

\begin{abstract}
Aiming at the problem that the current video anomaly detection cannot fully use the temporal information and ignore the diversity of normal behavior, an anomaly detection method is proposed to integrate the spatiotemporal information of pedestrians. Based on the convolutional autoencoder, the input frame is compressed and restored through the encoder and decoder. Anomaly detection is realized according to the difference between the output frame and the true value. In order to strengthen the characteristic information connection between continuous video frames, the residual temporal shift module and the residual channel attention module are introduced to improve the modeling ability of the network on temporal information and channel information, respectively. Due to the excessive generalization of convolutional neural networks, in the memory enhancement modules, the hopping connections of each codec layer are added to limit autoencoders' ability to represent abnormal frames too vigorously and improve the anomaly detection accuracy of the network. In addition, the objective function is modified by a feature discretization loss, which effectively distinguishes different normal behavior patterns. The experimental results on the CUHK Avenue and ShanghaiTech datasets show that the proposed method is superior to the current mainstream video anomaly detection methods while meeting the real-time requirements.
\end{abstract}


\IEEEpeerreviewmaketitle

\section{Introduction}
Video anomaly detection is a very challenging task in computer vision. At the same time, it also has important practical \cite{ref1} significance and has been widely used in traffic control, intelligent security, and social security. Video anomaly detection is the identification of objects and pedestrian behaviors that do not meet expectations in the video sequence, and in real life, there are three main difficulties \cite{ref2}. First of all, the probability of abnormal behavior is usually much smaller than the probability of normal behavior, and the positive and negative samples are uneven, and it is difficult to collect enough abnormal samples to train the model; Secondly, because abnormal behavior is usually unpredictable, resulting in a wide variety of abnormal samples, it is impossible to exhaust them; Finally, the definition of abnormal behavior may vary or even be opposite in different scenarios. For example, driving on the sidewalk is abnormal behavior, but driving on a motorway is normal.

Currently, the typical solution for video anomaly detection is to train an unsupervised learning model using a training set containing only normal samples and obtain the feature distribution of normal samples. If the model identifies video frames in the test set as outliers, they are regarded as abnormal frames \cite{ref3} containing abnormal behavior. According to the different methods of feature extraction, the methods of video anomaly detection are roughly divided into two categories: methods based on traditional machine learning and methods based on deep learning. The traditional machine learning method extracts histograms of gradients (HOG) \cite{ref4}, histograms of optical flow (HOF) \cite{ref5}, dense trajectories (DT) \cite{ref6} from videos, etc. Although this method has a simple network structure and strong interpretability, it will seriously affect the extraction of features under complex conditions, such as variable illumination and object occlusion, resulting in a significant decrease in detection accuracy.
\begin{figure*}
    \centering
    \includegraphics[width=0.95\textwidth]{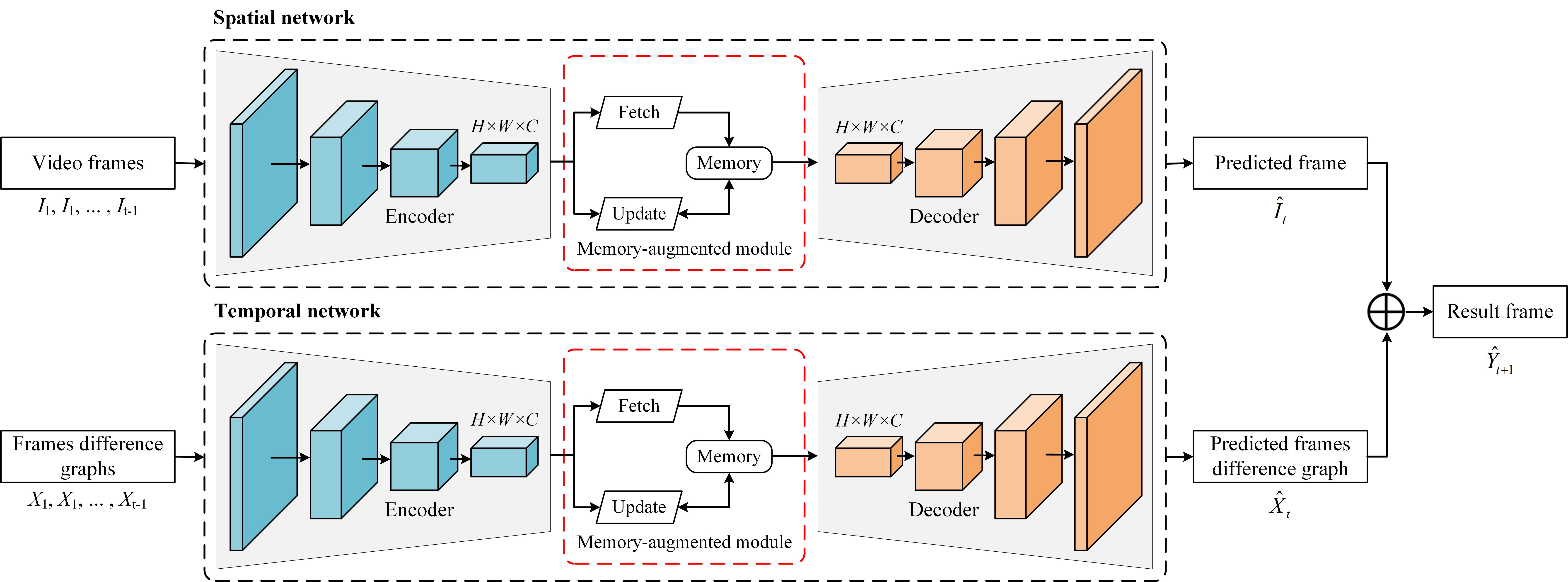}
    \caption{\centering{The network of video anomaly detection}}
    \label{fig1}
\end{figure*}

Among the methods based on deep learning, the reconstruction and prediction methods are two mainstream methods. The former usually adopts a framework based on auto-encoders (AEs) and generative adversarial networks (GANs). It uses anomalous samples to reconstruct assumptions with significant errors to detect anomalies, mainly including Conv-AE \cite{ref7}, ConvLSTM-AE \cite{ref8}, Conv3D-AE \cite{ref9}, DDGAN \cite{ref10}, etc. They outperform traditional methods by a large margin due to the ability to extract deeper behavioral features. However, its overly powerful representation ability will cause the reconstruction error of the abnormal sample to be as small as that of the normal sample, resulting in a high missed detection rate. The prediction rule considers the video's temporal characteristics, uses several consecutive frames to predict the next frame and uses the prediction error for abnormal discrimination. LIU et al. \cite{ref11} first proposed to predict future frames and add optical flow constraints to extract motion features. Compared to the reconstruction method, it has better performance, but at the cost of additional computation of the estimated optical flow between video frames and its vulnerability to pre-trained models. Some researchers have recently adopted hybrid methods to obtain better anomaly detection. TANG et al. \cite{ref12} concatenate two U-Net blocks in series, the former for predicting future frames and the latter for reconstructing frames. CHANG et al. \cite{ref13} use spatial autoencoders and temporal autoencoders to reconstruct individual frames and predict continuous RGB frame aberrations, respectively.

Inspired by the above research, this paper proposes a dual-stream network structure for separately predicting temporal and spatial information. Motion information between frames is captured using a residual temporal shift module (RTSM). The network enhances the extraction of key features through the residual channel attention module (RCAM). At the same time, the memory enhancement module is added to alleviate the problems caused by normal sample diversity and the strong generalization of convolutional neural networks (CNNs). Feature discretization loss is introduced to improve the model's ability to distinguish between different normal behavior patterns.

\section{Anomaly detection methods}
The video anomaly detection network architecture proposed in this article is shown in Fig. \ref{fig1}. The spatial (appearance) subnetwork is mainly used to extract the appearance features of pedestrians, and the temporal (action) subnetwork is mainly used to extract the movement characteristics of pedestrians. Both subnetworks share the same structure and consist of three main parts, including the encoder, decoder, and memory enhancement module, where the encoder and decoder form the autoencoder. In this paper, a continuous frame difference map is used instead of an optical flow map as the input of the temporal subnetwork because two pairs of adjacent video frames can directly subtract the frame difference map, and the network structure is simple and effective \cite{ref14}.

In the training stage, continuous video frames and continuous frame difference maps of the same length $t-1$ are first input into the spatial and temporal subnetwork. Then, after encoder encoding, reading the memory term in the memory enhancement module and decoding by the decoder, the predicted frame $\widehat{I}_\mathit{t}$ and the predicted frame difference graph $\widehat{X}_\mathit{t}$ are obtained. Finally, the results of the two subnetworks are added to obtain the final frame $\widehat{Y}_\mathit{t+1}$, and a normal model is learned using the objective function. During the testing phase, the test frame is determined to be abnormal by calculating the anomaly score.

\subsection{Autoencoder}
\begin{figure}
    \centering
    \includegraphics[width=0.4\textwidth]{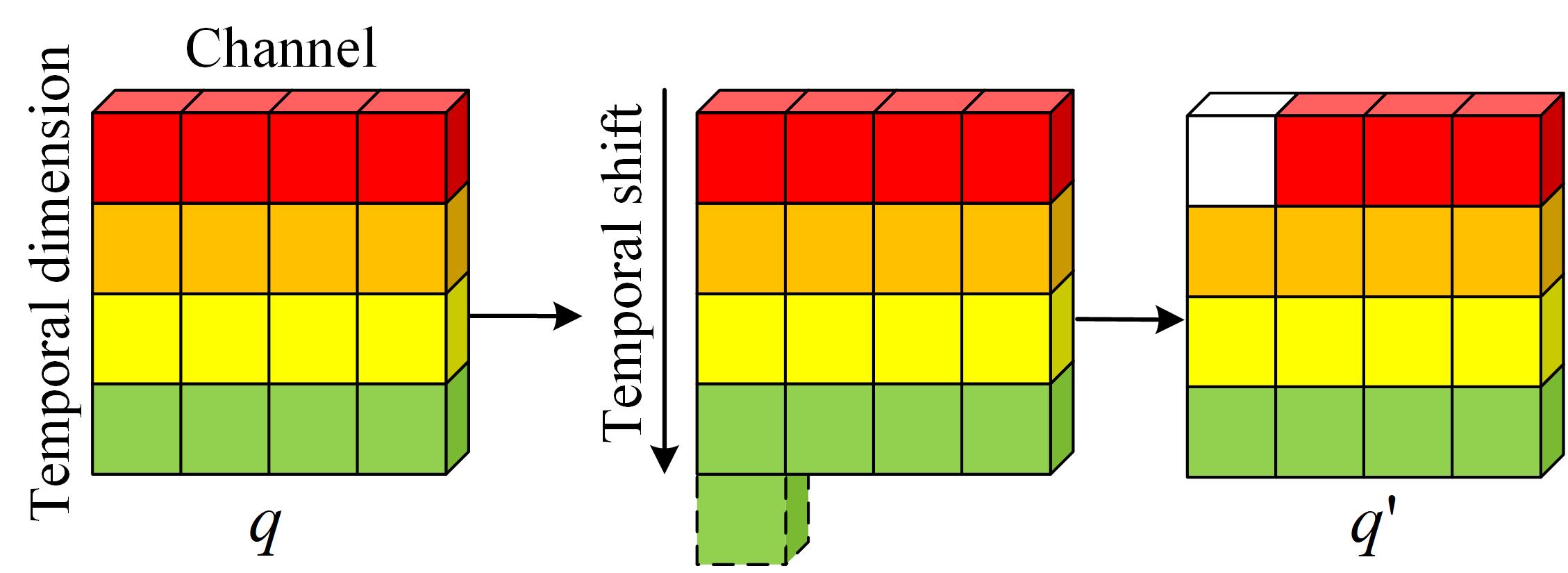}
    \caption{\centering{Temporal shift operation}}
    \label{fig2}
\end{figure}

Because 2D CNNs are challenging to capture the relationship between time dimensions and the computational cost of 3D CNNs is too high, to improve the network's modeling ability on temporal information, this paper introduces a temporal shift function block into the encoder. It cascades it with the 2D CNN to achieve the performance of 3D CNN while maintaining a relatively small amount of computation for 2D CNN \cite{ref15}. The temporal shift operation is performed along the time dimension $T$ of the feature map $\mathit{q}\in \mathbb{R}^{\mathit{T\times H\times W\times C}}$. While keeping the height $H$, width $W$, and channel number $C$ unchanged, a part of the current feature map is moved to the next frame. A part of the previous frame feature map is moved to the current frame so that part of the previous frame is combined with the features of the current frame to obtain the feature map $\mathit{q}^{\prime} $ after the temporal shift, as shown in Fig. \ref{fig2}. Moving part of the channels in the time dimension facilitates the information exchange between adjacent frames, and zero-computation and zero-parameter temporal modeling is achieved.

\begin{figure}
    \centering
	\includegraphics[width=0.5\linewidth]{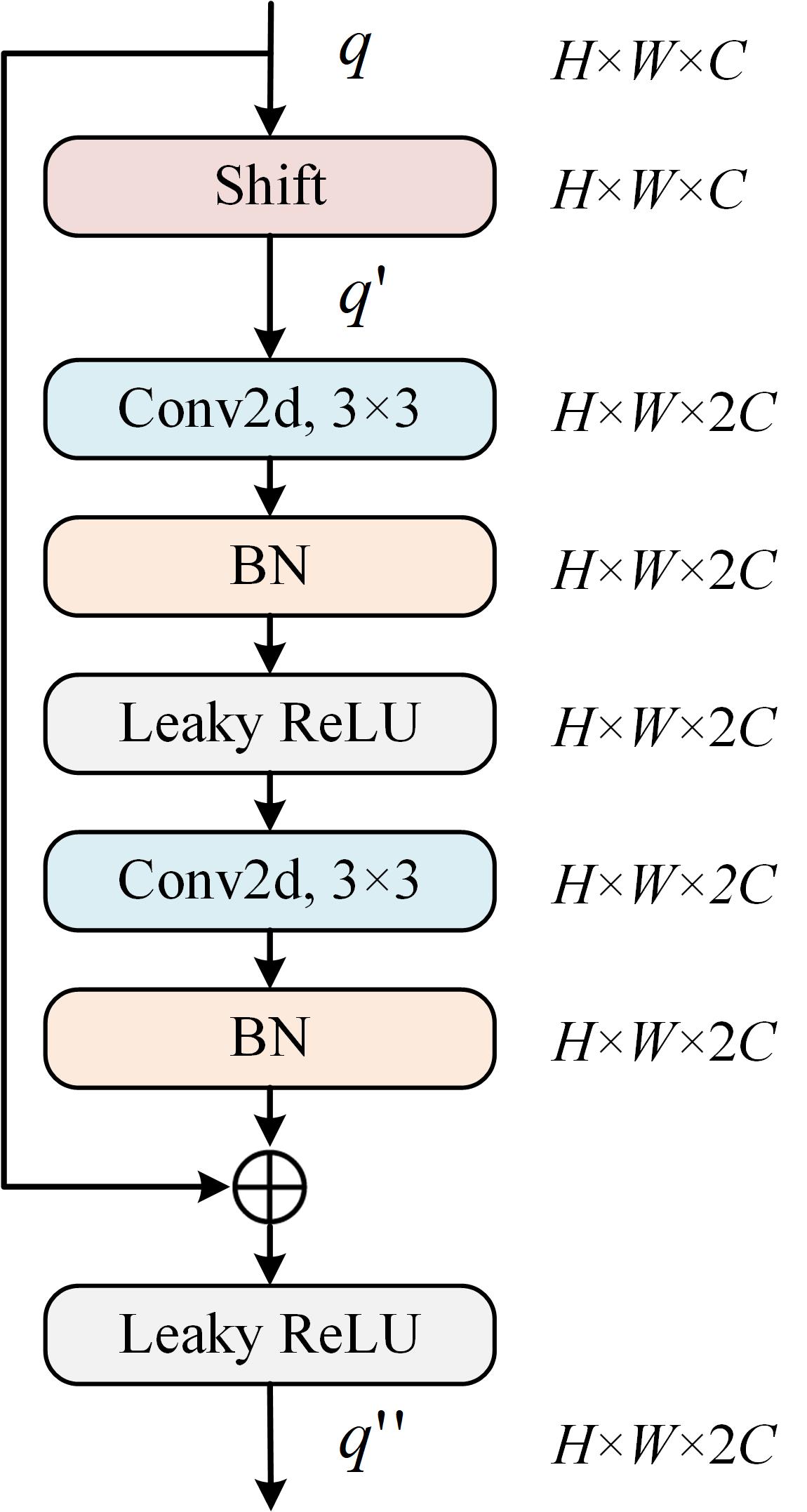}
	\caption{Residual temporal shift module}
	\label{fig3}
\end{figure}

In order to make the network learn temporal features without creating the problem of spatial feature learning degradation, this paper inserts the temporal shift function block into the interior of the residual branch instead of the outside, forming an RTSM to preserve the activation of the current frame, as shown in Fig. \ref{fig3}. The convolution operation is added after the temporal shift function block to enable the network to extract spatiotemporal features effectively.
\begin{equation}
\label{equation:1}
q^{\prime\prime}=\delta(q\oplus w_2\delta w_1(shift(q)))
\end{equation}

\noindent Where $shift(\bullet)$ represents temporal shift; $w_1$ and $w_2$ represent the weights of the two $3\times 3$ convolutional layers, respectively; $\delta$ represents the Leaky ReLU activation function; $\oplus$ denotes the addition of elements; $q^{\prime\prime}$ represents the new feature map.

In order to extract features that are more relevant to pedestrian behavior, RCAM is introduced into the decoder to improve the characterization capability of the network by making the network adaptively predict potential vital features, as shown in Fig. \ref{fig4}. This module modifies the characteristics of the channels by modeling the interdependencies between feature channels and applies them after each deconvolution layer. In RCAM, the channel attention block (CAB) is located after a convolutional block, which consists of $3\times 3$ convolutional layers, BN layers, Leaky ReLU layers, and $3\times 3$ convolutional layers cascading sequentially.
\begin{equation}
\label{equation:2}
U=W_2\delta(W_1q)
\end{equation}

Feature maps $\mathit{q}\in{\mathbb{R}^{\mathit{H\times H\times W\times C}}}$ are entered into the convolution block to obtain a new feature map $U$. Where $\delta$ represents the Leaky ReLU activation function; $W_1$ and $W_2$ represent the weights of the two $3\times3$ convolutional layers, respectively.

For CAB, to take advantage of channel dependence, firstly, $U$ is input into the global average pooling layer (GAP), and feature compression is performed along the spatial dimension. Each two-dimensional feature channel is converted into $C$ real numbers with global receptive fields, and $Z\in\mathbb{R}^{\mathit{1\times1\times C}}$ represents the output associated with the channel. Then the weights of each channel are learned through two $1\times 1$ convolutional layers, and $W_3$ is used to reduce the dimensionality of the feature channel to reduce the dimensionality ratio $r$ (set to 16) to capture the relationship between the channels fully. $W_4$ is used to restore the dimension of the feature channel.
\begin{equation}
\label{equation:3}
s(U)=\sigma(W_4\delta(W_3Z))
\end{equation}

\noindent Where $s(U)$ represents the weight of the corresponding channel; $\delta$ stands for Leaky ReLU activation function; $\sigma$ means Sigmoid activation function; $W_3$ and $W_4$ represent the weights of the two $1\times1$ convolutional layers, respectively.

Finally, $s(U)$ is regarded as the weight of each feature channel after selection \cite{ref16}, weighted to $U$ channel by multiplication, and the result is fused with $q$ through element addition to complete the correction of $q$ in the channel dimension.
\begin{equation}
\label{equation:4}
q^\prime=q\oplus(U\otimes s(U))
\end{equation}

\noindent Where $q\in \mathbb{R}^{\mathit{H\times W\times C}}$ and $q^\prime\in \mathbb{R}^{\mathit{H\times W\times C}}$ represent the input and output characteristic maps; $\otimes$ represents the product of elements; $\oplus$ indicates the addition of elements.

\subsection{Memory enhancement module}
In order to limit the overly powerful representation ability of CNNs, this paper refers to the method proposed by GONG et al. \cite{ref17} to introduce a memory enhancement module to learn a limited number of archetypal features that can best represent normal samples, which record various normal behavior patterns and store them in memory. The memory enhancement module will continuously read and update memory items alternately during the training phase. Firstly, the cosine similarity between each feature $q_t^k$ in the feature map ${q_t}\in \mathbb{R}^{\mathit{H\times W\times C}}$ and all memory terms $p_m$ is calculated to generate a 2-dimensional cosine similarity matrix of size $M\times K$. Here, $q_t^k$ represents the $k$ feature of the input feature map at time $t$, $p_m$ represents the $M$ memory term, $k(k=1, 2, 3, \dots, K, K=H\times W)$, and $m(m=1, 2, 3, \dots, M)$ represents the ordinal numbers of $K$ features and $M$ memory terms, respectively. $H$, $W$, and $C$ represent the feature map's height, width, and several channels. Then, apply the $Softmax$ function to the matrix in the direction of the $M$ dimension, and obtain the $K$ weights $w_t^{k,m}$ of each feature on all memory terms, as follows:
\begin{figure}
    \centering
	\includegraphics[width=0.5\linewidth]{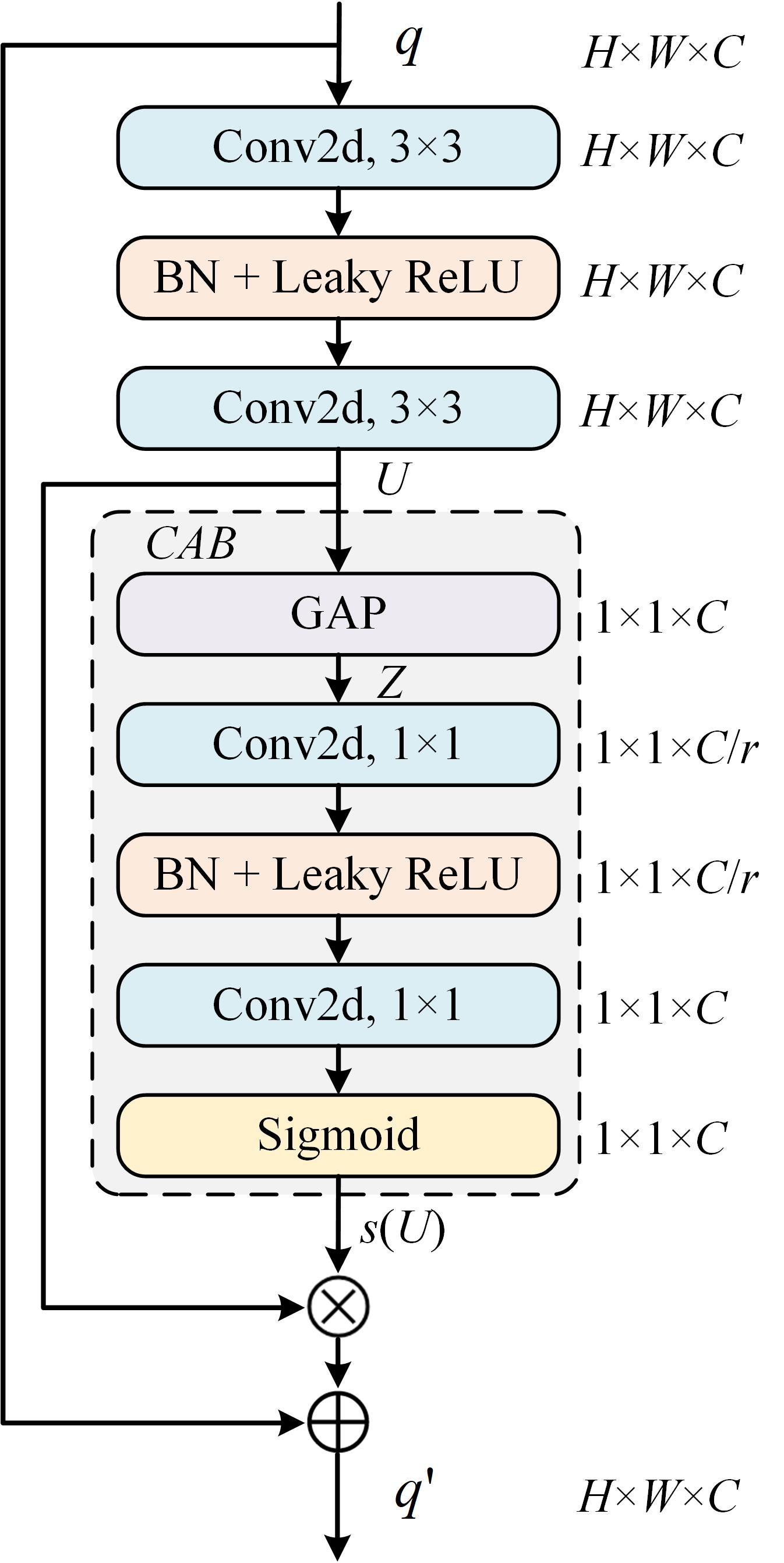}
	\caption{Residual channel attention module}
	\label{fig4}
\end{figure}

\begin{equation}
\label{equation:5}
w_t^{k,m}=\frac{\exp{((p_m)^Tq_t^k})}{\sum^{M}_{m=1}\exp{((p_m)^tq_t^k})}
\end{equation}

\begin{equation}
\label{equation:6}
\widehat{q}_t=\sum^{K}_{k=1}\sum^{M}_{m=1}w^{k,m}_tp_m 
\end{equation}

\begin{equation}
\label{equation:7}
v^{k,m}_t=\frac{\exp{((p_m)^Tq^k_t)}}{\sum^{K}_{k=1}\exp{((p_m)^Tq^k_t)}}
\end{equation}

Finally, for each $q_t^k$, a new feature $\widehat{q}^k_t$ is obtained by calculating the weighted sum corresponding to all memory terms. Repeat for $K$ features The above operation obtains a new feature map $\widehat{q}_t$ to realize the reading of memory terms. The update process is to use the extracted feature map to update the memory term, similar to Eq. (\ref{equation:5}), first apply the $Softmax$ function to the same two-dimensional cosine similarity matrix in the direction of $K$ dimension, and obtain the original weight $v^{k,m}_t$.

In this paper, the memory item with the most remarkable feature similarity is the feature's recent memory term. $U_t^m$ represents the index set of all features corresponding to the recent memory item. All the retrieved feature items with the most significant similarity to the memory item and the corresponding weight are indexed by $U_t^m$. Then normalize $v_t^{k,m}$ to get the new weights ${v^\prime}_t^{k,m}$. Finally, the weighted sum of the retrieved feature items is calculated to obtain the updated terms, and the original memory terms are further normalized after adding the updated terms. Repeat the above operation for $M$ memory items to obtain the updated memory term $\widehat{p}_m$. $f(\bullet)$ represents $L2$ normalization as follows:
\begin{equation}
\label{equation:8}
\widehat{p}_m=\sum^{M}_{m=1}f\left[p_m+\sum_{k\in U^m_t}v^{\prime k,m}_tq^k_t\right]
\end{equation}

\begin{figure*}
    \centering
    \includegraphics[width=0.9\textwidth]{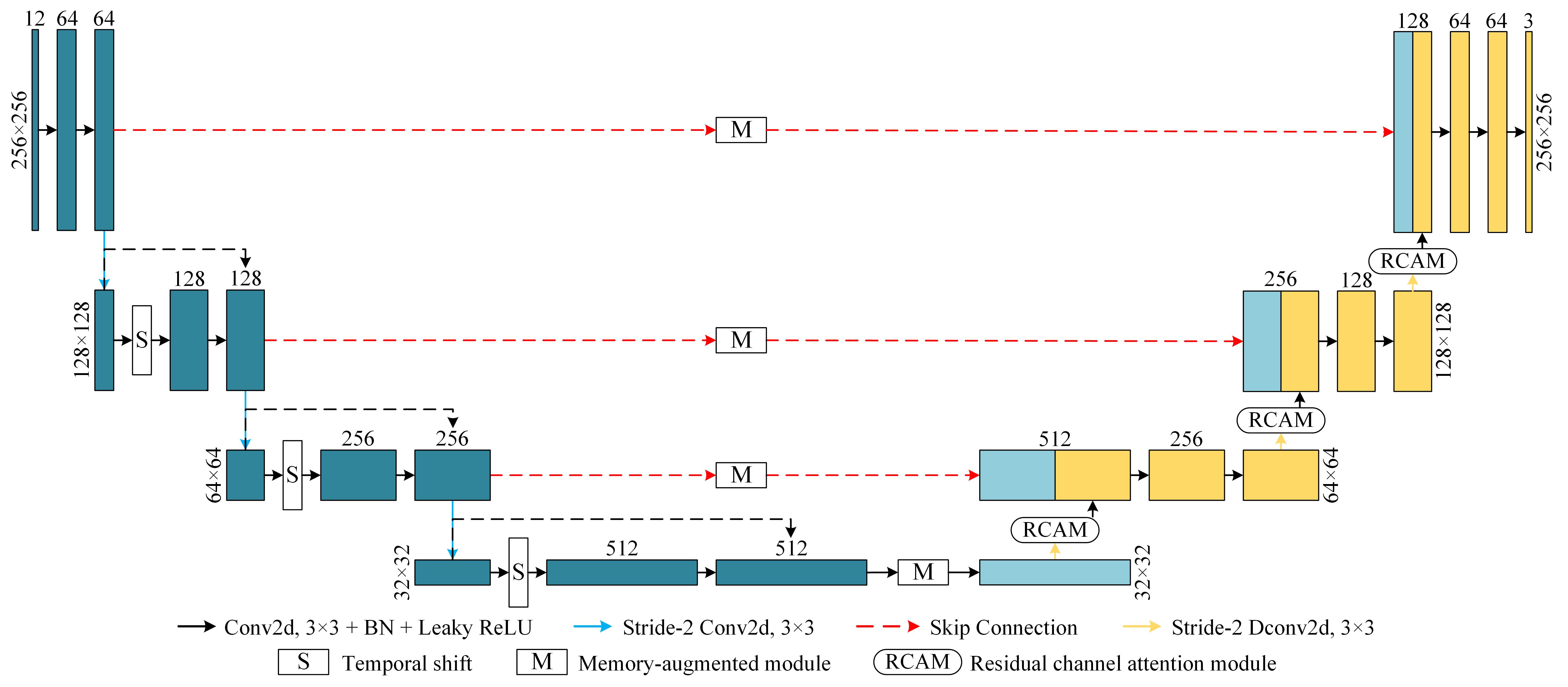}
    \caption{\centering{Improved spatial (temporal) sub-network}}
    \label{fig5}
\end{figure*}

Considering the diversity of normal behaviors, only adding a memory enhancement module at the bottleneck layer of the autoencoder cannot make the normal model learn enough normal behavior patterns. In this paper, skip connections are added between the feature layers of the exact resolution of the codec; simultaneously, a memory enhancement module is added in the middle of each layer of skip connections. In this way, the problem of information imbalance in each layer can be alleviated by fusing deep features and shallow features. The final frame, after decoding by the dual-stream network, is more similar to the normal frame by learning the prototype features of each scale of normal samples. When an anomaly frame is entered during the test phase, the difference between the final frame and it increases significantly, increasing the anomaly score of the anomaly frame and facilitating the detection of anomalous behavior. The improved spatial (temporal) subnet is shown in Fig. \ref{fig5}.

\subsection{Objective function}
The two subnetworks take the same loss function and define the objective function as the sum of the loss functions of the two subnetworks. Taking the spatial subnetwork as an example, the prediction loss uses the mean square difference loss function to minimize the $L2$ norm between the predicted frame $\widehat{I}$ and its true value $I$, as follows:
\begin{equation}
\label{equation:9}
\mathcal{L}_p=\|\widehat{I}-I\|^2_2
\end{equation}

In order to prevent the memory items from constantly tending to be the same as each other during the update process, this paper proposes a characteristic discrete loss $\mathcal{L}_s$ \cite{ref18}.
\begin{equation}
\label{equation:10}
\begin{split}
\mathcal{L}_s &= \sum^{T}_{t=1}\sum^{K}_{k=1}[(\|q^k_t-p_p\|^2_2 - \|q^k_t - p_{n1}\|^2_2 + a) \\
&+ (\|q^k_t - p_p\|^2_2 - \|p_{n2} - p_{n1}\| + b)]
\end{split}
\end{equation}

Among them, $p_p$ is the recent memory term of the characteristic $q_t^k$; $p_{n1}$ and $p_{n2}$ are the second and third nearest memory terms of $q_t^k$, respectively; $a$ represents the difference between the distance $p_{n1}$ and $q^k_t$ between $p_p$ and $q^k_t$; $b$ represents the difference between the distance between $p_p$ and $q^k_t$ and the distance between $p_{n1}$ and $p_{n2}$, set $a = 2$, $b = 1$.

While minimizing the distance between features and recent memory items, $\mathcal{L}_s$ maximizes the distance between the second and third nearest memory items so that the memory enhancement module tends to update the memory items closest to the features and achieves the purpose of separating all memory items, thereby enhancing the model's ability to distinguish between different normal behavior patterns. The loss function of the spatial subnetwork is balanced with a hyperparameter $\alpha_s$, as in Eq. (\ref{equation:11}).
\begin{equation}
\label{equation:11}
\mathcal{L}_i=\mathcal{L}_{p1}+\alpha_s\mathcal{L}_{s1}
\end{equation}

Similarly, the loss function of a temporal subnetwork is balanced with the hyperparameter $\beta_s$, as in Eq. (\ref{equation:12}).
\begin{equation}
\label{equation:12}
\mathcal{L}_x=\mathcal{L}_{p2}+\alpha_s\mathcal{L}_{s2}
\end{equation}

Considering that both subnetworks use the predictive method and use the same network structure, set $\alpha_s=\beta_s$. The objective function is balanced with the hyperparameters $\gamma_i$ and $\gamma_x$, set $\gamma_x=1-\gamma_i$ \cite{ref19}.
\begin{equation}
\label{equation:13}
\mathcal{L}=\gamma_i\mathcal{L}_i+\gamma_x\mathcal{L}_x
\end{equation}

\begin{figure*}[!t]
  \centering
    \subfigure[Test results on Avenue]{\includegraphics[width=0.45\textwidth]{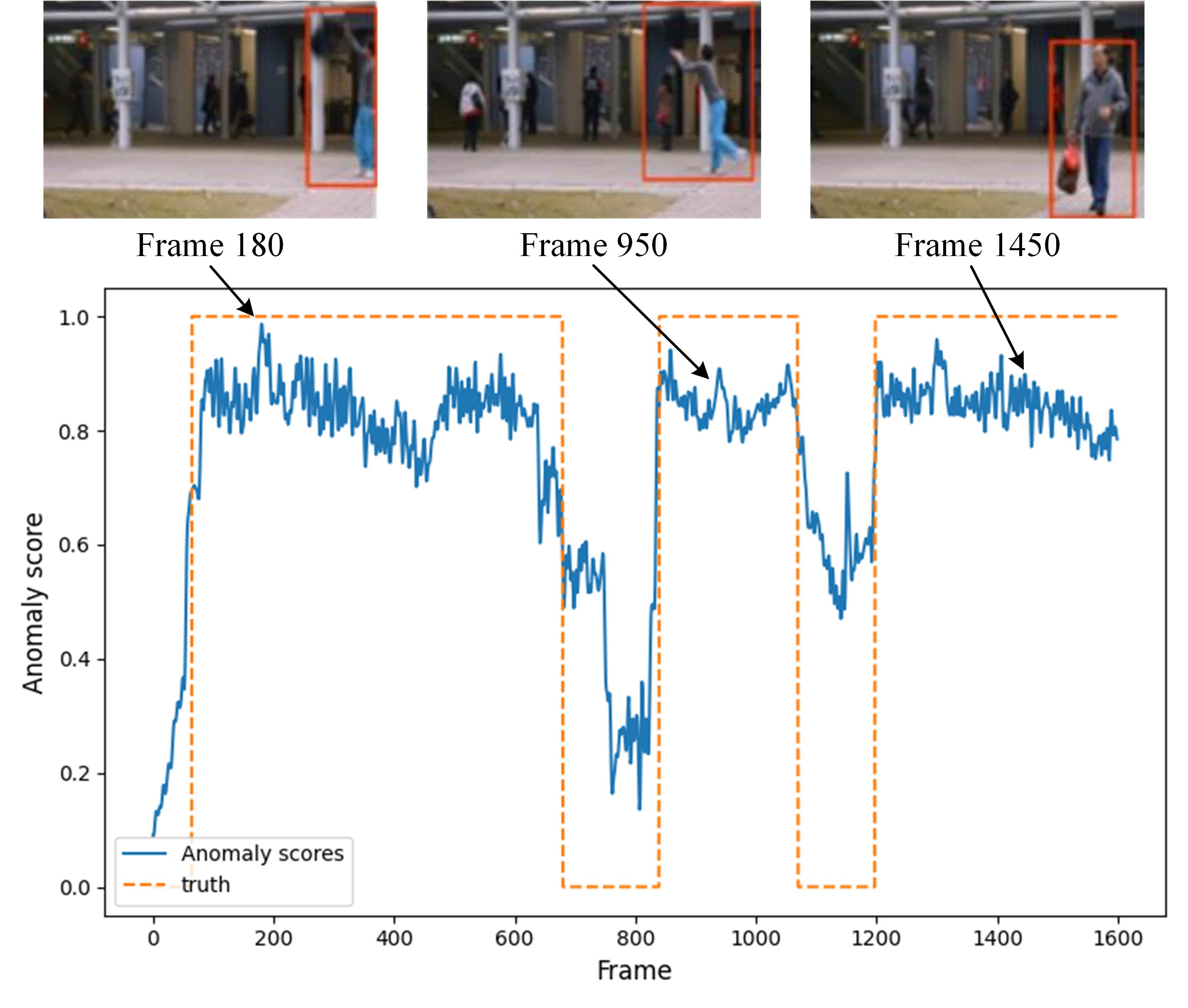}\label{fig6_a}} \qquad
    \subfigure[Test results on ShanghaiTech]{\includegraphics[width=0.45\textwidth]{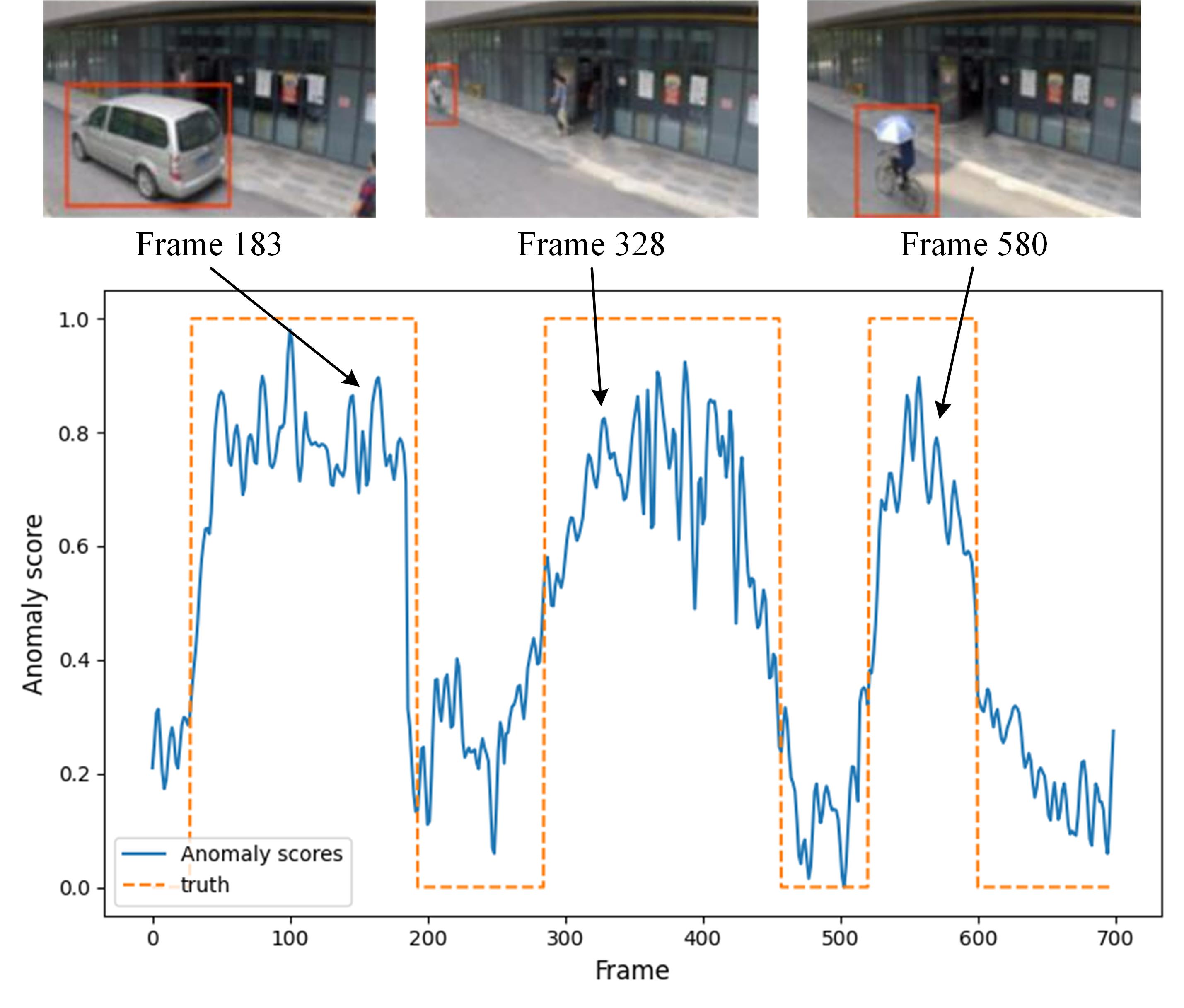}\label{fig6_b}}
    \caption{Test results on two datasets}
    \label{fig6}
\end{figure*}
\subsection{Anomaly score}
In Eq. (\ref{equation:14}), this paper uses the Peak Signal Noise Ratio (PSNR) between the test and final frames as an anomaly evaluation index. The smaller the PSNR value, the more severe the image distortion, and the more likely the test frame is to be abnormal.
\begin{equation}
\label{equation:14}
P(\widehat{Y}_{x,y,t},Y_{x,y,t})=10\log_{10}{\frac{\max(\widehat{Y}_{x,y,t})}{\frac{1}{N}\|\widehat{Y}_{x,y,t}-Y_{x,y,t}\|^2_2}}
\end{equation}

\noindent Where $Y_{x,y,t}$ represents the test frame; $\widehat{Y}_{x,y,t}$ represents the final frame; $x$ and $y$ represent the position of the pixel points of the frame image; $N$ represents the number of pixels in the frame graph; $\max(\bullet)$ represents the maximum pixel value of the framemap. Since the features extracted from the normal frame are similar to the memory term and the anomalous frame is the opposite, the degree of an anomaly in the test frame can be quantified by calculating the $L2$ norm between each feature and its most recent memory term, as in Eq. (\ref{equation:15}). According to this Eq. (\ref{equation:15}), $D_i(q_t,p)$ and $D_x(q_t,p)$ can be used to represent the anomaly scores of the temporal and spatial subnetworks, respectively. Therefore, the final anomaly score $S_t$ consists of 3 index fusions, balanced by the hyperparameter $\lambda$, where $g(\bullet)$ represents the minimum maximum normalization, as in Eq. (\ref{equation:16}).
\begin{equation}
\label{equation:15}
D(q_t,p)=\frac{1}{K}\sum^{K}_{k=1}\|q_t^k-p_p\|_2
\end{equation}

\begin{equation}
\label{equation:16}
\begin{split}
S_t &= \lambda(1-g(P(\widehat{Y}_{x,y,t},Y_{x,y,t})))+\frac{(1-\lambda)}{2}g(D_i(q_t,p)) \\
&+\frac{(1-\lambda)}{2}g(D_x(q_t,p))
\end{split}
\end{equation}

\begin{figure}[h]
  \centering
    \subfigure[Input frames]{\includegraphics[width=0.13\textwidth]{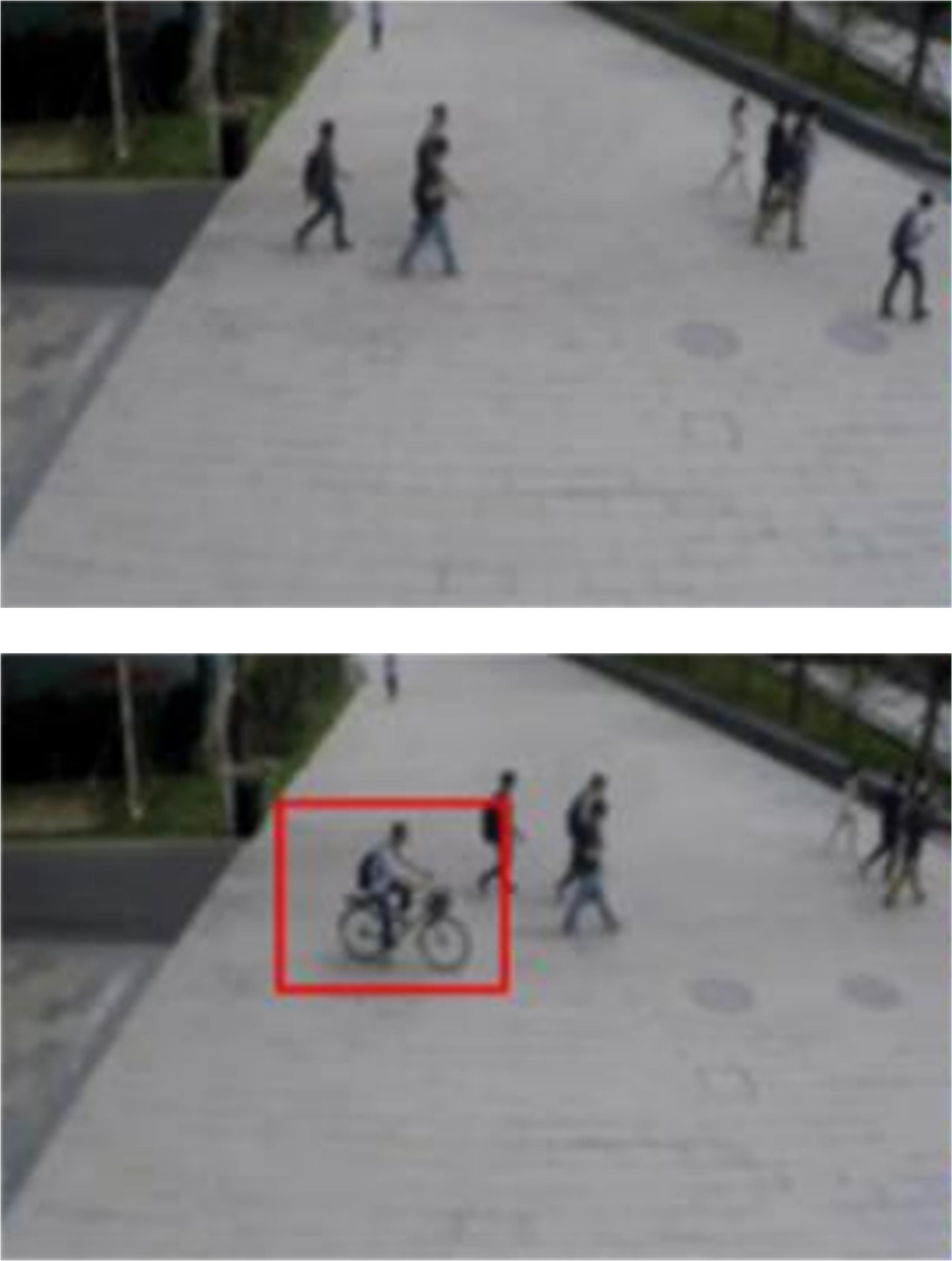}\label{fig7_a}} \quad
    \subfigure[Attention maps]{\includegraphics[width=0.13\textwidth]{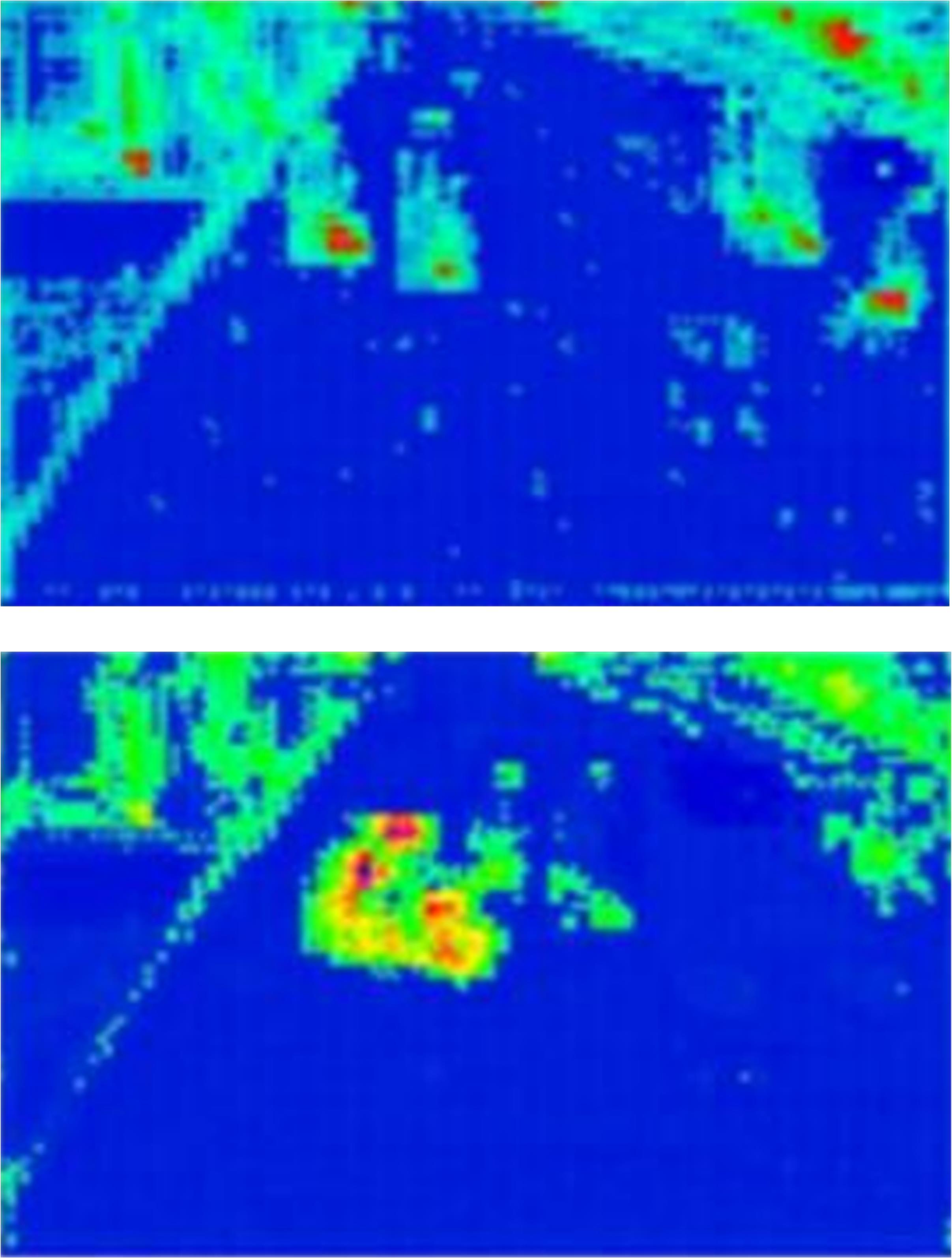}\label{fig7_b}} \quad
     \subfigure[Prediction error]{\includegraphics[width=0.13\textwidth]{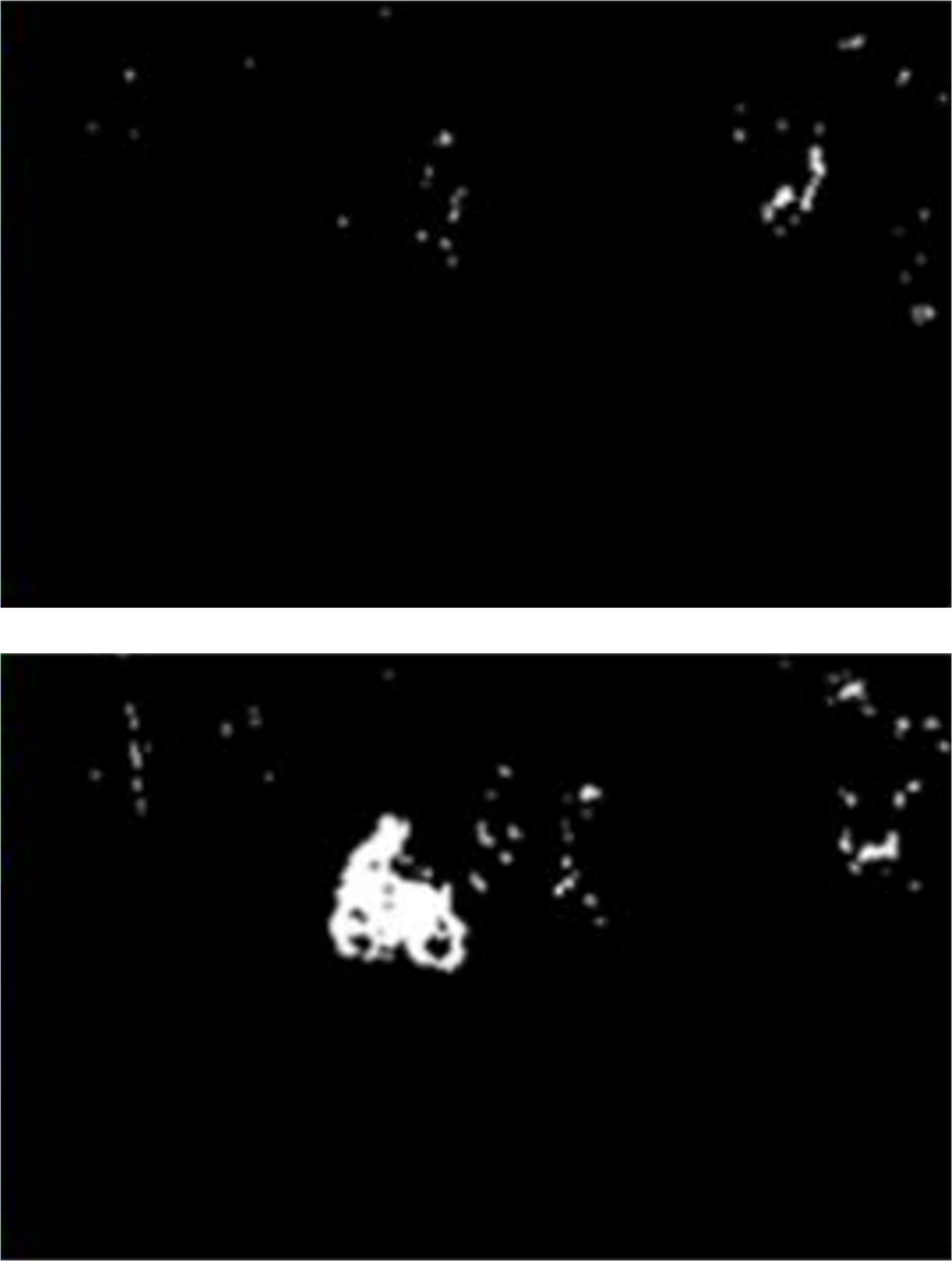}\label{fig7_c}}
    \caption{The visualization for ShanghaiTech}
    \label{fig7}
\end{figure}

\section{Experiments}
\subsection{Datasets and metrics}
Two standard datasets, CUHK Avenue \cite{ref20} and ShanghaiTech \cite{ref11}, were used in this experiment. The Avenue dataset was shot at a flat angle using a fixed camera parallel to the sidewalk, with a resolution of $360\times640$ video frames, 16 training videos (15328 frames) and 21 test videos (15324 frames), and abnormal behaviors including walking on grass, pedestrian jumping, loitering, and throwing objects. The ShanghaiTech dataset, shot using multiple fixed cameras on campus, has video frames with a resolution of $480\times856$ and contains 330 training videos (274515 frames) and 107 test videos (42,883 frames), with anomalous behaviors such as driving, cycling, and fighting. The dataset records 13 scenes with different lighting conditions and shooting angles, making it an extremely challenging data set.

In this experiment, the area under curve (AUC) curve of frame-level subjects was used as a quantitative evaluation index. Since AUC does not rely on artificially set thresholds, AUC is more objective as an evaluation indicator. The closer the AUC is to 1, the higher the anomaly detection accuracy and the better the model performance. In addition, frames per second (fps) are used as the detection speed metric to evaluate the temporal performance of the model.

\subsection{Experimental setup}
The server used in this experiment is equipped with GPU: NVIDIA Titan Xp, with 12G video memory; The CPU model is Intel Xeon E5-2620 v3@2.4GHz; 64-bit Ubuntu 16.04 operating system. The software environment uses the PyTorch framework and Python 3.6. The resolution of all input video frames is adjusted to $256\times256$, the pixel values are normalized to $[-1, 1]$, the number of memory items is set to $M=20$, and the length of the video sequence $t=5$. In the training phase, the network is trained using the Adam optimizer with momentum parameters $\rho_1=0.9$ and $\rho_2=0.999$ and $batch size=8$. The initial learning rate is set to 2e-5 and gradually reduced using cosine annealing. Referring to the relevant literature and experimental results, set the hyperparameters in the network as $\alpha_s=\beta_s=0.1$, $\gamma_i=\gamma_s=0.5$, $\lambda=0.8$. On the Avenue and ShanghaiTech datasets, the network trains 50 and 10 epochs, respectively, after loss convergence.

\begin{figure*}[t]
  \centering
    \subfigure[ROC curve on Avenue]{\includegraphics[width=0.45\textwidth]{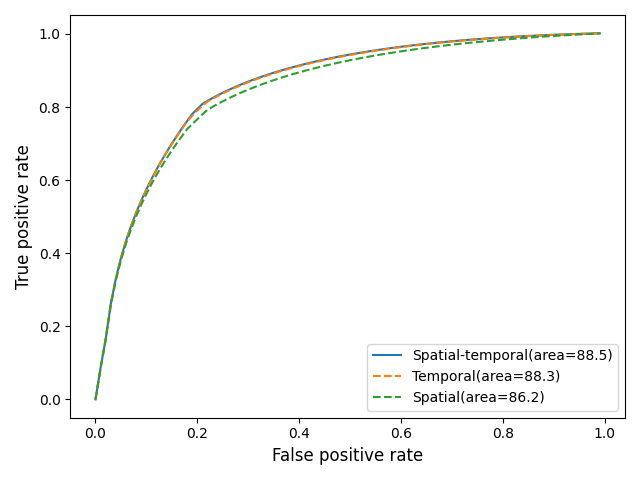}\label{fig8_a}} \quad
    \subfigure[ROC curve on ShanghaiTech]{\includegraphics[width=0.45\textwidth]{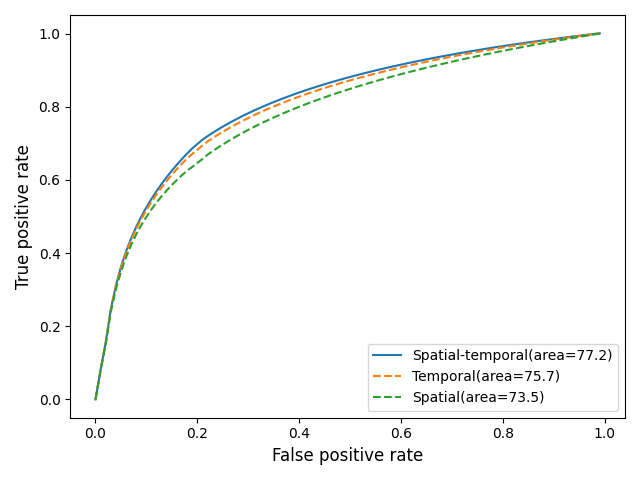}\label{fig8_b}}
    \caption{ROC curve on two datasets}
    \label{fig8}
\end{figure*}

\subsection{Experimental results and analysis}
\textbf{Qualitative experiments}. Fig. \ref{fig6_a} and \ref{fig6_b} show this method's approach's frame-level anomaly scoring curves on the Avenue and ShanghaiTech section test sets, respectively. Where the abscissa represents the number of frames, the ordinate represents the anomaly score, the solid line shows the change of the anomaly score over time, the dashed line represents the true value label for each frame, 0 represents the normal frame, and 1 represents the anomaly frame.

It can be seen from the Fig. \ref{fig6} that when there is an abnormal situation, the anomaly scoring curve increases rapidly, which indicates that the proposed method can detect abnormal behavior in the video in a timely and effective manner. There are three anomalous spikes in Fig. \ref{fig6_a}, caused by pedestrians throwing objects at frames 5480 and 6250 and pedestrians traveling in the wrong direction at frames 6750. Fig. \ref{fig6_b} also has three anomalous score spikes, with driving at frames 26083 and cycling at frames 26228 and 26480. According to different scenarios, times, and security requirements, an anomaly scoring threshold can be set to distinguish normal frames from abnormal frames, and timely alarms can be made for abnormal behaviors.

In order to visualize the anomalous behavior, this paper selects normal and abnormal frames in the ShanghaiTech dataset for visual comparison, as shown in Fig. \ref{fig7}. Fig. \ref{fig7_a} shows the true values of normal and abnormal frames; As can be seen from Fig. \ref{fig7_b}, the color around the cyclist is more prominent, accurately focusing on the area where abnormal behavior occurs; As you can see from Fig. \ref{fig7_c} when the pixels with a prediction error more significant than the average error within the frame are marked, the prediction error around the cyclist is greater than the prediction error around the pedestrian.

\begin{table}[!t]
\renewcommand{\arraystretch}{1.2}
\centering
\setlength{\tabcolsep}{5.5mm}
\caption{\label{table1}Average AUC for frame level detection} 
\resizebox{\linewidth}{!}
{
\begin{tabular}{ccc}
\hline
Method  &   Avenue   &   ShanghaiTech     \\
\hline
Conv-AE \cite{ref7}     &    80.0   &    60.9   \\
Mem-AE \cite{ref16}     &    83.3   &    71.2   \\
Pred\&Recon \cite{ref12}   &    83.7   &    71.5   \\
Frame-Pred \cite{ref11} &    85.1   &    72.8   \\
DDGAN \cite{ref10}  &    85.6   &    73.7   \\
AnoPCN \cite{ref21} &    86.2   &    73.6   \\
ROADMAP \cite{ref22} &    88.3   &    76.6   \\
Ours    &   \textbf{88.5}    &   \textbf{77.2}    \\
\hline
\end{tabular}}
\end{table}

\textbf{Quantitative experiments}. In order to further quantify the effectiveness of the proposed method, it was compared with the mainstream technical methods of recent years, and the results are shown in Table \ref{table1}.

As shown in Table \ref{table1}, the proposed method has the best results on the Avenue and ShanghaiTech datasets, with average AUCs of 88.5\% and 77.2\%, respectively. Because Avenue has a camera shake and the training set contains outliers, which affects network performance, the improvement is slight. Compared with the Conv-AE primary network, the method in this paper has been dramatically improved. Although Mem-AE also uses a memory enhancement module, the proposed method adds a memory enhancement module between the layers of the codec so that the decoder can fuse the prototype features of each level for prediction, so there is a significant improvement. Furthermore, very few memory items are used (20 items in this article compared to 2000 items for Mem-AE), reducing the memory space required. Compared with Pred\&Recon, this method uses two independent networks to train the model, and the results are added and fused without affecting each other, so the performance is better.

This paper tests the detection speed of the model using video frames of the exact resolution of $256\times256$. After experimental calculations, the proposed method requires an average of 0.029s to identify the anomaly of one frame. That is, the detection speed is about 34fps, and the detection speed of Mem-AE and Frame-Pred is 45fps and 25fps, respectively. Although the proposed method is slower than Frame-Pred's detection speed, the model detection performance is higher and meets the real-time requirements ($>$30fps).

To determine the hyperparameters in the objective function, the effects of different values of as, $\alpha_s$, $\beta_s$, $\gamma_x$ and $\gamma_i$ on AUC were experimented with based on relevant experience, and the results are shown in Table \ref{table2}.

As shown in Table \ref{table2}, when $\alpha_s$, $\beta_s$, $\gamma_i$, and $\gamma_x$ are set to 0.1, 0.1, 0.5, and 0.5, respectively, the model achieves the best results. The feature discretization loss is a correction to the prediction loss in the sub-network. When the values of $\alpha_s$ and $\beta_s$ are smaller, the value of AUC is higher. The loss function of the two sub-networks has roughly the same impact on the objective function, and the value of AUC is higher when the values of $\gamma_i$ and $\gamma_x$ are closer.
\begin{table}[!t]
\renewcommand{\arraystretch}{1.2}
\centering
\setlength{\tabcolsep}{5.5mm}
\caption{\label{table2}The influence of hyperparameters in the objective function on AUC}
\resizebox{\linewidth}{!}
{
\begin{tabular}{ccccc}
\hline
$\alpha_s$  &   $\beta_s$   &   $\gamma_i$  &   $\gamma_s$   & AUC(\%)   \\
\hline
0.2     &       0.2     &       0.4     &       0.6     &       87.0    \\
0.2     &       0.2     &       0.6     &       0.4     &       87.1    \\
0.2     &       0.2     &       0.5     &       0.5     &       88.0    \\
0.1     &       0.1     &       0.4     &       0.6     &       87.4    \\
0.1     &       0.1     &       0.6     &       0.4     &       87.5    \\
0.1     &       0.1     &       0.5     &       0.5     &       \textbf{88.5}    \\
\hline
\end{tabular}}
\end{table}

\subsection{Ablation Study}
To quantitatively verify the effectiveness of the proposed modules on network performance, ablation experiments were performed on the Avenue dataset, the results of which are shown in Table \ref{table3}.

As shown in Table \ref{table3}, the Baseline in this experiment is a traditional autoencoder structure composed of two-dimensional convolutions and only uses a predictive loss training network. The experimental results show that the memory enhancement module has an enormous improvement in network performance, RTSM and RCAM are roughly the same, and $\mathcal{L}_s$ is the smallest. The memory enhancement module forces the learning of a limited number of multi-scale prototype features. After directly entering them into the decoder, it will promote the prediction frame to tend to a normal frame, significantly increase the anomaly score of abnormal frames, and thus improve anomaly detection accuracy. The RTSM in the encoder and the RCAM in the decoder, respectively, strengthen the modeling ability of the network for temporal information and channel information and have improved to a certain extent in feature extraction. $\mathcal{L}_s$ only optimizes the training strategy, and the performance improvement is limited. The ablation experiment not only confirmed the effectiveness of each module but also proved the effectiveness of the overall framework.
\begin{table}[!t]
\renewcommand{\arraystretch}{1.2}
\centering
\setlength{\tabcolsep}{0.8mm}
\caption{\label{table3}Performance evaluation of each module of the network} 
\resizebox{\linewidth}{!}
{
\begin{tabular}{cccccc}
\hline
Method  &   RTSM   &   RCAM  &   Memory enhancement module  &   $\mathcal{L}_s$   & AUC($\%$)   \\ \hline
\multirow{2}{*}{Baseline} & $\times$   &  $\times$   &  $\times$   &  $\times$   &  78.4   \\
 & $\checkmark$   & $\times$   & $\times$   & $\times$   &  80.7    \\
\hline
\multirow{3}{*}{Ours}  & $\checkmark$   & $\checkmark$   &  $\times$   &  $\times$   &  83.2    \\ 
& $\checkmark$   &  $\checkmark$   & $\checkmark$   & $\times$   &  88.2    \\
& $\checkmark$   &  $\checkmark$   &    $\checkmark$   &  $\checkmark$   &  \textbf{88.5}    \\
\hline
\end{tabular}}
\end{table}

In order to verify the effectiveness of the spatiotemporal dual-stream network, a comparative experiment was designed on the Avenue and ShanghaiTech datasets, and the results are shown in Table \ref{table4}.

It can be seen from Table \ref{table4} that since the identification of abnormal pedestrian behavior is mainly based on motion information, the performance of the temporal subnetwork based on motion characteristics is better than that of the spatial subnetwork. At the same time, the performance of spatiotemporal dual-stream networks is higher than that of both, indicating that the proposed method makes full use of information in time and space. Fig. \ref{fig8_a} and \ref{fig8_b} represent the ROC curves of the model on the Avenue and ShanghaiTech datasets, respectively, with the abscissa being the false positive rate (FPR), the ordinate being the true positive rate (TPR), and the area under the curve being the average AUC. The solid line represents the spatiotemporal dual-stream network, the dashed line represents the temporal subnetwork, and the dotted line represents the spatial subnetwork.

\begin{table}[!t]
\renewcommand{\arraystretch}{1.2}
\centering
\setlength{\tabcolsep}{3.5mm}
\caption{\label{table4}Comparison of single-stream network and dual-stream network}
\resizebox{\linewidth}{!}
{
\begin{tabular}{ccc}
\hline
Method  &   Avenue      &    ShanghaiTech   \\ 
\hline
spatial subnetwork      &       86.2    &   73.5    \\
temporal subnetwork     &       88.3    &   75.7    \\
spatiotemporal dual-stream network      &       \textbf{88.5}   &   \textbf{77.2}   \\
\hline
\end{tabular}}
\end{table} 

\section{Conclusions}
This paper proposes a video anomaly detection method based on pedestrian spatio-temporal information fusion, which predicts the continuous frame difference map and continuous video frames through the spatio-temporal dual-stream network and uses the final fused frame for anomaly detection. The improved convolutional autoencoder introduces RTSM, RCAM, and memory enhancement modules, which can extract rich pedestrian spatiotemporal features and key feature representations and increase the diversity of learned prototype features and ease the generalization of CNN substantial questions. At the same time, discrete feature loss is added to improve the model's ability to distinguish various normal behaviors. The experimental results on two standard data sets show that the method in this paper has achieved good detection accuracy and speed. Moreover, the effectiveness of the improved module and the spatio-temporal dual-stream network is verified by ablation experiments. During the experiment, it was found that background noise explicitly impacts the model's performance. Future work plans to design a network that is more robust to background noise to improve abnormal behavior's detection accuracy further.

\end{document}